\documentclass[letterpaper, 10 pt, conference]{ieeeconf}  %

\IEEEoverridecommandlockouts                              %

\usepackage{times}

\usepackage{multicol}
\usepackage[bookmarks=true]{hyperref}

\usepackage{amsmath}
\usepackage{amssymb}
\usepackage{amsfonts}
\usepackage{graphicx}
\usepackage{color}
\usepackage{algorithm}
\usepackage{algpseudocode}

\newcommand\blfootnote[1]{%
  \begingroup
  \renewcommand\thefootnote{}\footnote{#1}%
  \addtocounter{footnote}{-1}%
  \endgroup
}

\begin{document}

\title{PI-ARS: Accelerating Evolution-Learned Visual-Locomotion with\\ Predictive Information Representations}

\author{Kuang-Huei Lee$^{*1}$ \quad Ofir Nachum$^{*1}$ \quad Tingnan Zhang$^{1}$ \quad Sergio Guadarrama$^{1}$ \quad Jie Tan$^{1}$ \quad Wenhao Yu$^{1}$%
\thanks{*Equal contribution}%
\thanks{$^{1}$All authors are with Google Research, \newline 1600 Amphitheatre Parkway Mountain View, CA 94043, United States
        \{\tt\small {leekh, ofirnachum, tingnan, sguada, jietan, magicmelon\}@google.com}}%
}

\maketitle

\begin{abstract}

Evolution Strategy (ES) algorithms have shown promising results in training complex robotic control policies due to their massive parallelism capability, simple implementation, effective parameter-space exploration, and fast training time. However, a key limitation of ES is its scalability to large capacity models, including modern neural network architectures. In this work, we develop Predictive Information Augmented Random Search (PI-ARS) to mitigate this limitation by leveraging recent advancements in representation learning to reduce the parameter search space for ES. Namely, PI-ARS combines a gradient-based representation learning technique, Predictive Information (PI), with a gradient-free ES algorithm, Augmented Random Search (ARS), to train policies that can process complex robot sensory inputs and handle highly nonlinear robot dynamics. We evaluate PI-ARS on a set of challenging visual-locomotion tasks where a quadruped robot needs to walk on uneven stepping stones, quincuncial piles, and moving platforms, as well as to complete an indoor navigation task. Across all tasks, PI-ARS demonstrates significantly better learning efficiency and performance compared to the ARS baseline. We further validate our algorithm by demonstrating that the learned policies can successfully transfer to a real quadruped robot, for example, achieving a 100\% success rate on the real-world stepping stone environment, dramatically improving prior results achieving 40\% success.\blfootnote{The supplementary video is available at \href{https://kuanghuei.github.io/piars}{\color{magenta}kuanghuei.github.io/piars}} %

\end{abstract}

\IEEEpeerreviewmaketitle

\begin{figure*}[t]
\centering
\includegraphics[width=0.65\linewidth]{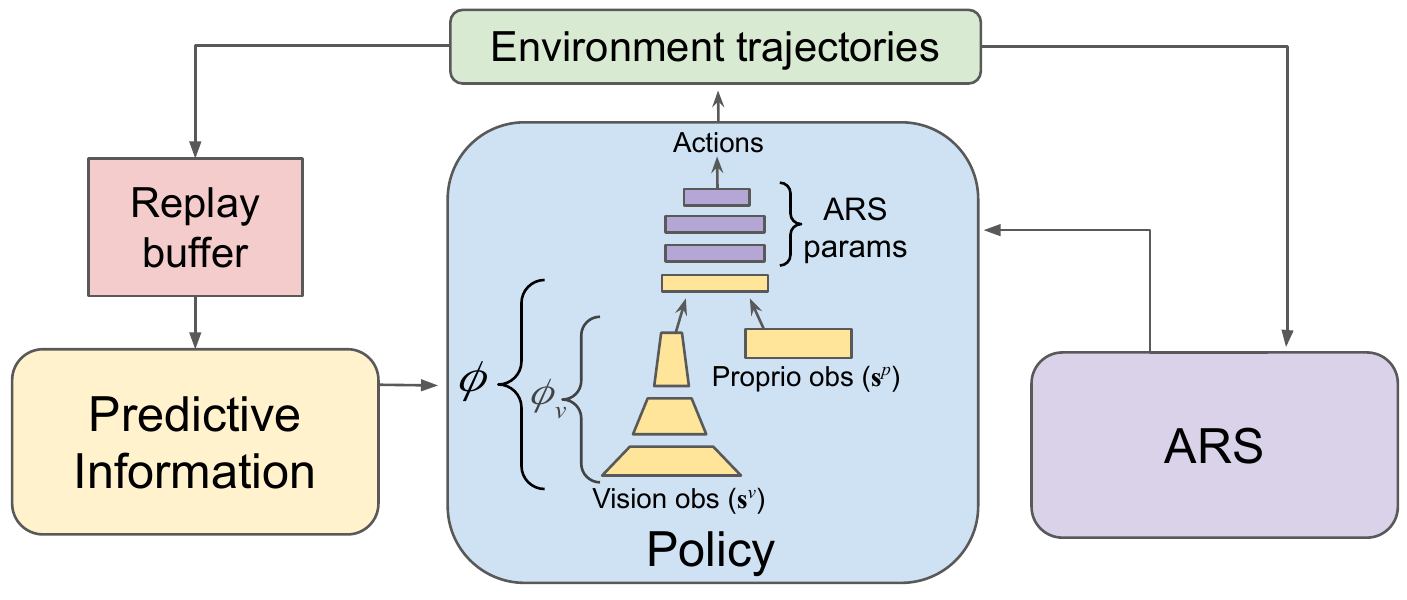}
\caption{In PI-ARS, an observation encoder $\phi$ is used to provide a compressed observation representation. This way, ARS learning is focused on a much smaller neural network than if the whole policy (mapping robot observations to actions) were learned end-to-end. As ARS uses sampled trajectories from the policy to update its parameters, PI-ARS uses the same trajectories to maintain a replay buffer for training $\phi$ via predictive information (PI), which includes auxiliary-learned networks as described in Section~\ref{subsec:pi}.
}
\label{fig:piars}
\end{figure*}

\section{Introduction}

Evolution Strategy (ES) optimization techniques have received increasing interest in recent years within the robotics and deep reinforcement learning (DRL) communities \cite{salimans2017evolution, mania2018simple, yu2021visual, song2020rapidly, yu2020learning, cully2015robots, huang2020accelerated}. 
ES algorithms have been shown to be competitive alternatives to commonly used gradient-based DRL algorithms such as PPO~\cite{schulman2017proximal} and SAC~\cite{haarnoja2018soft}, while also 
enjoying the benefits of massive parallelism, simple implementation, effective parameter-space exploration, and faster training time~\cite{salimans2017evolution}. 

Despite the promising progress of ES algorithms, they nevertheless exhibit key limitations when compared to gradient-based DRL algorithms. 
Namely, unlike gradient-based methods, ES methods scale poorly to high-dimensional search spaces, commonly encountered when using high-capacity modern neural network architectures~\cite{nesterov2017random,salimans2017evolution}.
An important and exemplary task is visual-locomotion \cite{yu2021visual}, in which a legged robot relies on its vision input to decide where to precisely place its feet to navigate uneven terrains. Due to the rich and diverse sensor observations as well as complex robot dynamics, learning such a task requires the use of deep convolutional neural networks (CNNs) with a large number of learnable parameters, thus exacerbating the sample-complexity of ES methods. 

In this paper, we develop Predictive Information Augmented Random Search (PI-ARS) to relieve this key bottleneck of ES algorithms. 
Our key insight is to leverage the power of gradient-based and gradient-free learning together by modularizing the learning agent into two components: (1) an encoder network mapping high-dimensional and diverse observation inputs to a concise fixed-length vector \emph{representation}, and (2) a smaller policy network that maps the compressed representations to actions. 
For (1), we leverage the power of gradient-based learning, and use a self-supervised objective based on maximizing the \emph{predictive information} (PI) of the output representation, inspired by previous work in representation learning \cite{oord2018representation,ha2018world,lee2020predictive,srinivas2020curl,yang2021representation,chen2021empirical}.
Meanwhile for (2), we leverage the simplicity and parallelizability of the ES optimization method Augmented Random Search (ARS) \cite{mania2018simple}.
By decoupling representation learning from policy optimization in this way, we avoid scalability issues while fully leveraging the advantages of ES methods.

We evaluate our proposed PI-ARS algorithm on a variety of visual-locomotion tasks, both in simulation and on a quadruped robot. Among these tasks, the robot is evaluated on its ability to walk on uneven stepping stones, quincuncial piles, and moving platforms, as well as to complete an indoor navigation task (Figure \ref{fig:sim_task_suite}). Through extensive experimentation in simulation, we find that PI-ARS significantly outperforms the baselines (ARS \cite{yu2021visual}, SAC \cite{haarnoja2018soft}, PI-SAC \cite{lee2020predictive}), both in training speed and final performance. We further validate the results by deploying the learned policies on a real quadruped robot. Using the same physical setup as prior work \cite{yu2021visual}, PI-ARS learns more robust policies that can consistently finish the entire course of stepping stones, achieving 100\% success over 10 real-robot trials, compared to 40\% success rate of prior work. We observe similarly successful robustness to real-world transfer for the indoor navigation policy. 

In summary, the contributions of this paper are the following:
\begin{enumerate}
    \item We propose a new PI-ARS algorithm that combines the advantages of gradient-based self-supervised representation learning and gradient-free policy learning, thus solving a key bottleneck of ES algorithms.
    \item We apply PI-ARS in visual-locomotion tasks, which significantly improve the state-of-the-art \cite{yu2021visual} both in simulation and in the real world.
\end{enumerate}

\section{Related Work}

\subsection{Evolution Strategy for RL}

There have been numerous works that demonstrate the effectiveness of applying Evolution Strategy (ES) algorithms to continuous control problems \cite{Tan:2014, salimans2017evolution, mania2018simple}. For example, Tan et al. applied Neural Evolution of Augmenting Topologies (NEAT) to optimize a character to perform bicycle stunts \cite{Tan:2014}. Within the field of deep reinforcement learning, Salimans et al. first demonstrated that ES algorithm can be applied to train successful neural-network policies for the OpenAI Gym control tasks \cite{salimans2017evolution}. Mania et al. introduced Augmented Random Search (ARS), a simple yet effective ES algorithm that further improves the learning efficiency for robotic control tasks \cite{mania2018simple}. 

Compared to gradient-based RL algorithms, ES algorithms can handle non-differentiable dynamics and objective functions, explore effectively with sparse or delayed rewards, and are naturally parallelizable. As such, researchers have applied ES algorithms in a variety of applications such as legged locomotion \cite{yu2021visual, jain2019hierarchical}, power grid control \cite{huang2020accelerated}, and mixed autonomy traffic \cite{vinitsky2018benchmarks}. However, as ES algorithms do not leverage backpropagation, they suffer from low sample efficiency and may not scale well to complex high-dimensional problems \cite{salimans2017evolution}. As a result, applying ES to learn vision-based robotic control policies is rarely explored \cite{yu2021visual}. 

\subsection{Predictive Representations for RL}
Our work relies on learning representations that are predictive of future events. 
Prior work has shown benefit from having good models of the past and future states \cite{schmidhuber1990making, schmidhuber1991reinforcement, schmidhuber2015learning}.
More recently, using these principles to guide state representation learning methods has been demonstrated to yield favorable performance both in practice~\cite{lee2020predictive,yang2021representation} and in theory~\cite{nachum2021provable,yang2021trail}. 
A natural approach to learning such representations is using generative models to explicitly predict observations \cite{ha2018world, hafner2019dream}, which could be challenging and expensive for high-dimensional tasks.
Alternatively, using variational forms~\cite{poole2019variational} of the predictive information \cite{bialek1999predictive}, commonly leading to contrastive objectives, makes learning such representations more tractable \cite{lee2020predictive, yang2021representation, oord2018representation}.
In this work, we take the contrastive approach to learn predictive representations of the observed states, upon which we learn an optimal policy with augmented random search (ARS) \cite{mania2018simple}, an ES method.
To the best of our knowledge, the proposed learning system is the first to take such a combined approach, and the first to apply predictive information representations on visual locomotion tasks with legged robots.

\subsection{Visual-Locomotion for Legged Robots}

Visual-locomotion is an important research direction that has received much attention in the robotics research community \cite{mastalli2017trajectory, magana2019fast, villarreal2020mpc, fankhauser2018robust, jenelten2020perceptive, grandia2020multi, gangapurwala2020rloc, miki2022learning, kim2020vision}. Directly training a visual-locomotion controller is challenging due to the high dimensional visual input and the highly nonlinear dynamics of legged robots \cite{yu2021visual, margolis2021learning}. Many existing methods manually and explicitly decouple the problem into more manageable components including perception \cite{fankhauser2018robust, kim2020vision}, motion planning \cite{jenelten2020perceptive, park2015online}, and whole body control \cite{dicarlo2018dynamic, kim2019highly, bledt2017policy}. 
In this work, we consider direct learning of visual-locomotion controllers for quadruped robots as the test-bed for our proposed learning algorithm and demonstrate that by combining gradient-free ES and gradient-based representation learning techniques, we can enable more effective and efficient learning of visual-locomotion controllers.

\section{Method}

In this section we describe our method, PI-ARS, which combines representation learning based on predictive information (PI), and Augmented Random Search (ARS)~\cite{mania2018simple}.
See Figure~\ref{fig:piars} for a diagram of the algorithm and Algorithm~\ref{alg:piars} for a pseudocode.

\subsection{Problem Formulation and Notations}
PI-ARS solves a sequential decision-making environment in which at each timestep $t$ the agent is presented with an \emph{observation} $\mathbf{s}_t$.
In visual-locomotion, this observation typically includes a visual input $\mathbf{s}^v_t$ (e.g., depth camera images) as well as proprioceptive states $\mathbf{s}^p_t$.
The agent's \emph{policy} determines its behavior, by providing a mapping from observations $\mathbf{s}_t$ to \emph{actions} $\mathbf{a}_t$. 
After application of action $\mathbf{a}_t$, the agent receives a reward $r_t$ and a new observation $\mathbf{s}_{t+1}$. 
This process is repeated until the agent is terminated, either due to a timeout or encountering a terminal condition (e.g., the robot falls). 
The agent's \emph{return} is computed as the sum of rewards over an entire episode, and the agent's goal is to maximize the expected value of the return.

\subsection{Predictive Information}
\label{subsec:pi}
A good observation encoder for policy learning must provide representations that are both \emph{compressive} -- so that ARS learning is focused on much fewer parameters than learning from raw observations would entail -- and \emph{task-relevant} -- so that ARS has access to all features necessary for learning optimal behavior.
To this end, we propose to learn an encoder $\phi$ to maximize predictive information (PI).
In general, PI refers to the mutual information between past and future $I(\mathrm{past}; \mathrm{future})$~\cite{bialek1999predictive}.
In our setting involving environment-produced sub-trajectories $\tau=(\mathbf{s}_t,\mathbf{a}_t,r_t,\mathbf{a}_{t+1},r_{t+1},\dots,\mathbf{a}_{t+k-1},r_{t+k-1}, \mathbf{s}_{t+k})$, $\mathrm{past}$ corresponds to $(\mathbf{s}_t,\mathbf{a}_t,\dots,\mathbf{a}_{t+k-1})$ and $\mathrm{future}$ refers to the both the per-step rewards $(r_t,\dots,r_{t+k-1})$ and the ultimate visual observation $\mathbf{s}_{t+k}^v$\footnote{This empirical choice of $\mathrm{future}$ works well in our setting.}; i.e., $\mathbf{s}_{t+j} = (\mathbf{s}^v_{t+j}, \mathbf{s}^p_{t+j})$.

We use $\phi$ to map both $\mathbf{s}_t$ and $\mathbf{s}_{t+k}^v$ to a lower dimensional representation. 
Namely, as shown in Figure \ref{fig:piars}, the observation encoder $\phi$ contains a vision encoder $\phi_v$ that maps visual observations $\mathbf{s}^v_t$ to a 128-d representation. 
This representation is subsequently concatenated with proprioceptive states, and the concatenation is projected to be the output of $\phi$, which is also 128-d.
We thus use the entire encoder $\phi$ to encode $\mathbf{s}_t$, while use only the vision encoder $\phi_v$ to encode the future $\mathbf{s}_{t+k}^v$. %
By learning $\phi$ to maximize the mutual information between these $\mathrm{past}$ and $\mathrm{future}$, we ensure that $\phi$ encodes the necessary information in $\mathbf{s}_t$ to predict the future. 
Notably, previous work has shown that representations that are predictive of the future are also provably beneficial for solving the task~\cite{nachum2021provable}.

To learn $\phi$, we use a combination of two objectives, the first corresponding to reward prediction and the second to $\mathbf{s}_{t+k}^v$ prediction.\footnote{We note that in our early experiments, we found learning $\phi$ using reward prediction alone leads to insufficient representations for solving the tasks.}
For the former, we simply predict rewards $\hat{r}_t,\dots,\hat{r}_{t+k-1}$ using an RNN over inputs $(\phi(\mathbf{s}_t),\mathbf{a}_t,\dots,\mathbf{a}_{t+k-1})$. 
At time $t+j$, the RNN cell takes a latent state and an action and outputs a reward prediction $\hat{r}_{t+j}$ and the next latent state.
The reward loss is defined as
\begin{align}
\small
    L_{r} = \mathbb{E}_{\tau}\bigg[{\sum_{j=0}^{k-1} (\hat{r}_{t+j} - r_{t+j})^2}\bigg].
\label{eq:reward_pred}
\end{align}
This corresponds to maximizing $I(\mathrm{past}; \mathrm{future\ rewards})$ with a generative model \cite{fischer2020conditional}.

For the prediction of $\mathbf{s}_{t+k}^v$, rather than a generative model, as used for $\hat{r}_{t+j}$, which would present challenges for high-dimensional image observations, 
we leverage InfoNCE, a contrastive variational bound on mutual information~\cite{oord2018representation,poole2019variational,lee2020predictive}.
We use an auxiliary learned, scalar-valued function $f$ to estimate the mutual information using access to samples of sub-trajectories $\tau=(\mathbf{s}_t,\mathbf{a}_t,\dots,\mathbf{a}_{t+k-1}, \mathbf{s}_{t+k})$.
Specifically, we conveniently exclude the course of actions and choose the following form:
\begin{small}
\begin{multline}
    I(\mathrm{past}; \mathrm{future\ obs}) \geq \tilde{I}(\mathrm{past}; \mathrm{future\ obs}) = \\  ~\mathbb{E}_{\tau}\bigg[f(\phi(\mathbf{s}_t),\phi_v(\mathbf{s}^v_{t+k})) \\
    - \log\mathbb{E}_{\tilde{\mathbf{s}}_{t+k}}[\exp\{ f(\phi(\mathbf{s}_t),\phi_v(\tilde{\mathbf{s}}^v_{t+k}))\}]\bigg],
\label{eq:pi_infonce}
\end{multline}
\end{small}where $\tilde{\mathbf{s}}^v_{t+k}$ is from an observation randomly sampled independent of $\tau$. Our objective is then to maximize this variational form with respect to both $\phi$ and $f$. %

To parameterize $f$, we use an MLP to map $\phi(\mathbf{s}_t)$ to $\mathbf{z}^{\mathrm{past}}_{t}$, a 128-d vector. Meanwhile, we map $\phi_v(\mathbf{s}^v_{t+k})$ to $\mathbf{z}^{\mathrm{future}}_{t+k}$, a 128-d vector representation of the future, using another MLP.
The function $f$ is then computed as a scalar dot-product of the two vectors.

We train both the reward objective and the variational objective using batch samples from a replay buffer of sub-trajectories collected by ARS. %
To approximate $\mathbb{E}_{\tilde{\mathbf{s}}_{t+k}}$ in Equation~\eqref{eq:pi_infonce}, we use samples of $\tilde{\mathbf{s}}^v_{t+k}$ within the same batch of sub-trajectories. 
The full objective is optimized using the Adam stochastic gradient descent optimizer~\cite{kingma2014adam} and the gradient is calculated using back-propagation. 
Full implementation details are included in the Appendix.

\begin{algorithm}
\caption{Pseudocode for PI-ARS.}\label{alg:cap}
\begin{algorithmic}
\State Initialize encoder $\phi$, auxiliary networks (RNN, $f$), and optimizer $\mathrm{Opt}_{\phi,\mathrm{aux}}$.
\State Initialize ARS parameters $\theta$ and optimizer $\mathrm{Opt}_\theta$. 
\State Initialize replay buffer $\mathcal{B}$.
\For{$T=1,\dots$}
\State \textit{\#\#\#\# ARS \#\#\#\#}
\State Sample $\{\sigma_i\}_{i=1}^N$ from normal with scale $\delta$.
\State Collect environment trajectories for policies $(\phi,\theta \pm \sigma_i)$.
\State Compute returns $R_{i,\pm}$ for each policy.
\State Compute gradient of $M$ best-performing directions:
\State $~~~~\hat{g} = \frac{\delta}{M} \sum_{i=1}^M (R_{i,+} - R_{i,-}) \sigma_i$.
\State Update $\theta$ w.r.t. gradient $\hat{g}$ and $\mathrm{Opt}_\theta$.
\State Add trajectories to $\mathcal{B}$.
\State \textit{\#\#\#\# PI \#\#\#\#}
\State Sample batch of sub-trajectories $\{\tau_i\}_{i=1}^B$ from $\mathcal{B}$.
\For{$i=1,\dots,B$}
\State $\hat{L}_i=-\tilde{I}(\tau_i) + L_r(\tau_i)$ \hfill Equations \eqref{eq:reward_pred}, \eqref{eq:pi_infonce}
\EndFor
\State Compute total loss $\hat{L} = \sum_{i=1}^B \hat{L}_i$.
\State Update $\phi$ and aux. networks w.r.t. loss $\hat{L}$ and $\mathrm{Opt}_{\phi,\mathrm{aux}}$.
\EndFor
\end{algorithmic}
\label{alg:piars}
\end{algorithm}

\begin{figure*}[tb]
\centering
\includegraphics[width=0.99\linewidth]{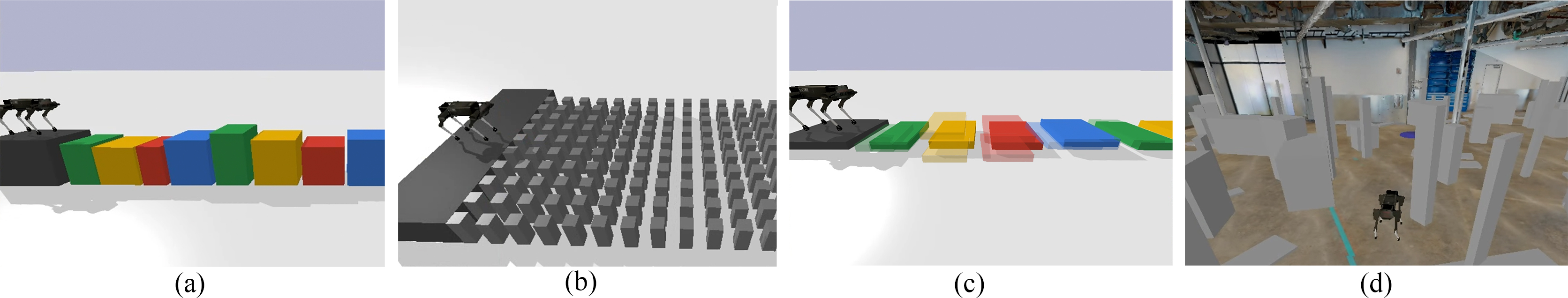}
\caption{The environments that we used to benchmark the PI-ARS learning system: (a) uneven stepping stones, (b) quincuncial piles, (c) moving platforms (the afterimage is for indicating that these platforms are moving), and (d) indoor navigation.}
\label{fig:sim_task_suite}
\end{figure*}

\begin{figure*}[tb]
\centering
\includegraphics[width=0.95\linewidth]{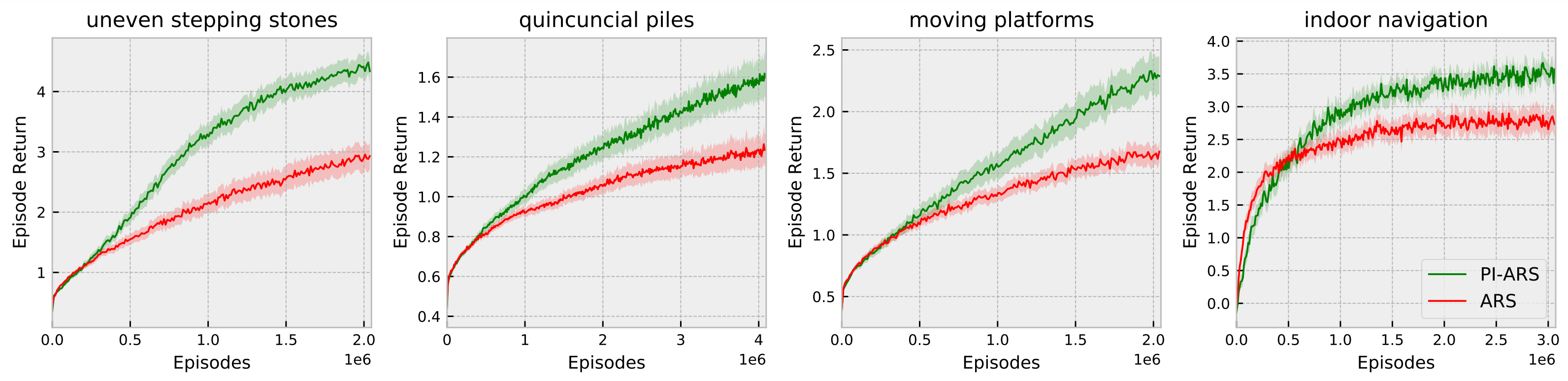}
\caption{Simulation results. We compare the performance of PI-ARS to ARS during training on four challenging simulation environments. PI-ARS consistently and significantly outperforms ARS.}
\label{fig:sim}
\end{figure*}

\subsection{PI-ARS}
\label{subsec:piars}
The encoder $\phi$ learned by PI maps a high-dimensional observation to a concise 128-d representation, upon which we use ARS to train a more compact policy network as follows.
At each iteration of ARS, the algorithm samples $N$ perturbations $[\sigma_1, \dots, \sigma_N]$ of the policy weights $\theta$ using a standard normal distribution with scale $\sigma$. The algorithm evaluates the policy returns at $\theta+\sigma_i$ and $\theta-\sigma_i$. 
ARS then computes an estimation of the policy gradient by aggregating the returns from the best-performing perturbation directions: 
\begin{equation}
\hat{g} = \frac{\delta}{M} \sum_{i=1}^M (R_{i,+} - R_{i, -}) \sigma_i,
\end{equation}
where $\delta$ is the update coefficient, $M$ is the number of top-performing perturbations to be considered, and $R_{i,+},R_{i,-}$ denote the total return of the policy at perturbations $\theta\pm\sigma_i$. 
We refer the readers to \cite{mania2018simple} for additional details.

We iterate between updating the representation network $\phi$ with the PI objective and updating the policy with ARS. 
To maximize data re-use, we store the sampled trajectories from perturbed policies evaluated by ARS in a replay buffer used for the PI learning pipeline.

\section{Experiments}

We aim to answer the following questions in our experiments: 
\begin{itemize}
    \item Is our proposed algorithm, PI-ARS, able to learn vision-based policies that solve challenging visual-locomotion tasks?
    \item Does PI-ARS achieve better performance than alternative methods that do not apply representation learning?
    \item Are our learned policies applicable to real robots?
\end{itemize}

\subsection{Visual-Locomotion Tasks}

To answer the above questions, we design a variety of challenging visual locomotion tasks.
Figure \ref{fig:sim_task_suite} shows the suite of environments that we evaluate on. More details of each environment can be found in Section \ref{sec:supp_env}.

\paragraph{Uneven stepping stones}
In this task, the robot is tasked to walk over a series of randomly placed stepping stones separated by gaps, and elevation of the stepping stones changes dramatically (Figure \ref{fig:sim_task_suite} (a)).

\paragraph{Quincuncial piles}
This is an extension to uneven stepping stones, where we reduce the contact surface area and arrange stones in both forward and lateral directions (Figure \ref{fig:sim_task_suite}(b)).

\paragraph{Moving platforms}

We construct a set of stepping stones and allow each piece to periodically move either horizontally and vertically at a random speed (Figure \ref{fig:sim_task_suite}(c)).

\paragraph{Indoor navigation with obstacles}
In this task, we evaluate the performance of PI-ARS controlling the robot to navigate in a cluttered indoor environment (Figure \ref{fig:sim_task_suite} (d). Specifically, we randomly place boxes on the floor of a scanned indoor environment and command the robot to walk to a target position. 

\subsection{Experiment Setup}
\label{subsec:experiment_setup}

We use the Unitree Laikago quadruped robot \cite{laikagounitree}, which weighs $24$kg and has $12$ actuated joints, with two depth cameras installed: one Intel D435 in the front for a wider field of view and one Intel L515 on the belly for better close-range depth quality. 
We create a corresponding simulated Laikago robot in the PyBullet physics simulator \cite{coumans2017pybullet} with physical properties from hardware spec and simulated cameras that matches the camera intrinsics and extrinsics from the real cameras. 
The observation, action, and reward designs are detailed as follows.

\subsubsection{Observation Space}
We design the observation space in our visual-locomotion tasks following prior work by Yu et al.~\cite{yu2021visual}. 
In particular, our observation space consists of two parts:
$\mathbf{s}=(\mathbf{s}^v, \mathbf{s}^p$), where $\mathbf{s}^v$ are the two $32\times24$ images from depth sensors, and $\mathbf{s}^p$ include all the proprioceptive states (and controller states). 
In our experiments, $\mathbf{s}^p = (\mathbf{q}_s, \dot{\mathbf{p}}, \dot{\Phi}, \dot{\Theta}, \mathbf{r}_{1\ldots4}, c_{1\ldots4}, \phi_{1\ldots4}, \mathbf{a}_{prev})$ includes the CoM height, roll, and pitch $\mathbf{q}_s = (p_z, \Phi,\Theta)$, the estimated CoM velocity $\dot{\mathbf{p}}$, the gyroscope readings $\dot{\Phi}, \dot{\Theta}$, the robot's feet positions $\mathbf{r}_{1\ldots4}$ in the base frame, the feet contact states $c_{1\ldots4}$, the phase of each leg in its respective gait cycle $\phi$, and the previous action.

For the indoor navigation task, we additionally include the relative goal vector $\mathbf{n} = \mathbf{\bar{o}} - \mathbf{p}$ as part of the observation, where $\mathbf{\bar{o}}$ is the target location and $\mathbf{p}$ is the robot's position.

\subsubsection{Action Space}
\label{ssec:action_space}
We follow the prior work \cite{yu2021visual} to use a hierarchical design for the visual-locomotion controller with a trainable high-level vision policy $\pi_\theta$ that maps visual and proprioceptive input to a high-level motion command, and an MPC-based low-level motion controller that executes the high-level motion command with trajectory optimization. 
The high-level motion command, i.e. the action space for the RL problem, is defined as: $(\mathbf{q}_s^d, \mathbf{\dot{q}}^d, \mathbf{r}^d_{1\ldots4}, \mathbf{h}_{1\ldots4})$, where $\mathbf{q}_s^d$ and $\mathbf{\dot{q}}^d$ are the desired CoM pose, velocity, $\mathbf{r}^d_{i}$ is $i$th foot's target landing position $(r_{xi}, r_{yi}, r_{zi})$, and $\mathbf{h}_{i}$ is the peak height of $i$th foot's swing trajectory.

\subsubsection{Reward Function}
For training a policy to walk on different terrains, we use the following reward function:
\begin{equation}
R(\mathbf{s}, \mathbf{a}) = clip(\dot{p}_x, -\dot{p}_{x}^{max}, \dot{p}_{x}^{max}) - w |\Psi|,
\end{equation}
where $\dot{p}_x$ is the CoM velocity in the forward direction, and $\Psi$ the base yaw angle. The first term rewards the robot to move forward with a maximum speed controlled by $\dot{p}_{x}^{max}$, the second term encourages the robot to walk straightly.

For the indoor navigation task, we use the delta geodesic (path) distance to the goal as our reward:

\begin{equation}
R(\mathbf{s}, \mathbf{a}, \mathbf{\bar{o}}) = d_g^t - d_g^{t-1},
\end{equation}
where $d_g^t$ is the geodesic distance between the robot and the target location at time $t$. 

\subsubsection{Early termination.} A training episode is terminated if: 1) the robot loses balance (CoM height $p_z$ below 0.15~m, pitch $|\Theta| > 1$~rad, or roll $|\Phi| > 0.3$~rad in our experiments), or 2) the robot reaches an invalid joint configuration, e.g. knee bending backwards.

\subsection{Learning in Simulation}
\label{subsec:sim_results}

In this subsection, we discuss the results of PI-ARS learned on simulated visual-locomotion tasks and compare to a state-of-the-art ARS approach to robotic visual-locomotion \cite{mania2018simple,yu2021visual} (Figure~\ref{fig:sim}).
Among other baseline approaches that we have tried include SAC \cite{haarnoja2018soft} and PI-SAC \cite{lee2020predictive} but both algorithms failed to make any non-negligible learning progress for the tasks we consider despite extensive hyperparameter tuning, and so we omit these algorithms from the results. 
For fair comparison, all algorithms utilize the same MPC-based locomotion controller and learn policies in the high-level command space described in Section~\ref{ssec:action_space}.
All policies utilize the same network architecture; i.e., the policy learned by the baseline ARS method is composed of the same set of convolution and feed-forward layers used for $\phi$ in PI-ARS.

We train PI-ARS and ARS policies using a distributed implementation. %
For all PI-ARS and ARS experiments, we perform $N=1024$ perturbations per ARS iteration and use the top 50\% performers ($M=512$) to update the policy network head.
This choice, determined through a grid search, empirically works the best for both PI-ARS and ARS in our implementation.
Further increase of $N$ (and thus computation cost) does not significantly improve the performance.
The algorithm is run until convergence with a maximum of $4000$ training iterations, resulting in a maximum of $2,048,000$ simulation episodes per trial.
We perform $30$ trials of training PI-ARS/ARS with uniformly sampled $\sigma$ and $\delta$ values ($\sigma \sim [0.005, 0.05]$, $\delta \sim [0.005, 0.05]$) and report the mean and standard error of returns against number of training episodes for each task.

As we demonstrate in the supplementary video, PI-ARS is able to learn vision-based policies that successfully solve these challenging visual-locomotion tasks.
Figure~\ref{fig:sim} shows that on all tasks, PI-ARS gives significantly better returns and sample-efficiency than the ARS baseline.
For example, on uneven stepping stones, the mean return after 2,000,000 episodes of training improves by 48.01\%, from 2.93 to 4.34.
This empirically demonstrates the effectiveness of learning ARS policies upon compressed, gradient-learned representations instead of end-to-end.
On the other hand, observing that SAC fails to learn, we hypothesize that the advantages of ARS such as parameter-space exploration and stability are critical to these complex visual-locomotion tasks.
Furthermore, adding predictive information to SAC, i.e. PI-SAC, does not improve learning, suggesting that even with an effective representation learner, without a powerful policy solver, a learning algorithm is not able to sufficiently tackle these visual-locomotion tasks.

\begin{figure*}[tb]
\centering
\includegraphics[width=0.95\linewidth]{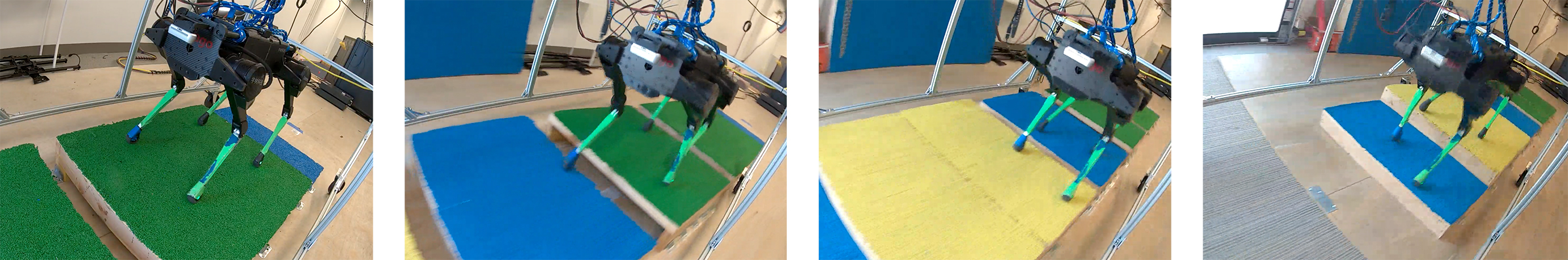}
\caption{PI-ARS policy solving a challenging real-world visual-locomotion task involving a series of four step stones separated by gaps. PI-ARS successfully completes this terrain, avoiding all gaps, 100\% of the time measured over 10 trials.}
\label{fig:real_exp}
\end{figure*}

\begin{figure}[tb]
\centering
\includegraphics[width=\linewidth]{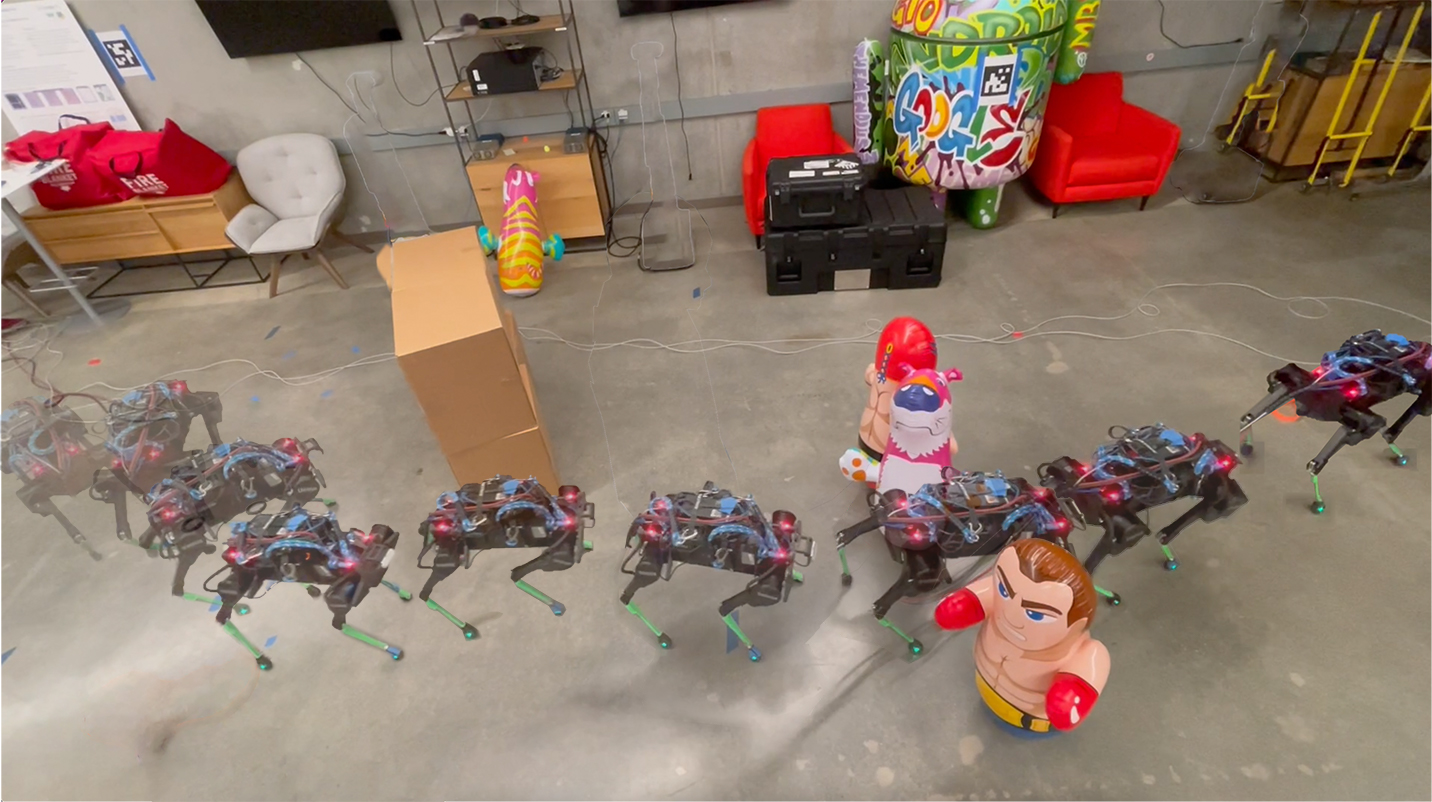}
\caption{PI-ARS policy learns to navigate in a cluttered real-world indoor environment.}
\label{fig:navigation}
\end{figure}

\subsection{Validation on Real Robot}
We deploy the visual-locomotion policy trained in simulation on a Laikago robot to perform two visual-locomotion tasks: 1) walking over real-world stepping stones (Figure~\ref{fig:real_exp}), and 2) navigating in an indoor environment with obstacles (Figure~\ref{fig:navigation}).

To overcome the sim-to-real gap, we adopt the same procedure as done by Yu et al. \cite{yu2021visual}. 
For the visual gap, during training, we first apply random noise to the simulated depth images to mimic the real-world depth noises.
Then we apply a Navier-Stokes-based in-painting operation \cite{bertalmio2001navier} with radius of 1 to fill the missing pixels, followed by down-sampling to $32 x 24$ (with OpenCV's $\mathrm{INTER\_AREA}$ interpolation method for resizing~\cite{opencv_library}). %
On the real hardware, we obtain $640 x 480$ raw depth images from both L-515 and D435 cameras and perform the same in-painting and down-sampling. 
To mitigate the dynamics gap, we apply dynamics randomization during training.

Videos of our real-world experiments can be found in the supplementary material.

\paragraph{Stepping Stones} For the stepping stones task, we created a physical setup consisting of four stones separated by three gaps between $[0.12, 0.18]$m (Figure \ref{fig:real_exp}). The PI-ARS policy is learned in simulation with an easier version of uneven stepping stones where stone heights change less significantly.
Our PI-ARS policy was able to achieve $100\%$ success rate on the stepping stone environment with $10$ trials. 
In contrast, the ARS baseline \cite{yu2021visual} with the same training and evaluation setting achieved $40\%$ success rate for reaching the last stone with all four legs and often failed at the last gap. 

\paragraph{Indoor Navigation} For evaluating the navigation task in the real world, we design a route in an indoor environment with obstacles (Figure \ref{fig:navigation}). The robot needs to navigate to the target location while avoiding the obstacles. To enable the robot to better avoid the obstacles, we rotate the front camera of the robot such that it can see $\sim$3 meters ahead of the robot. We also track the robot base position using a motion capture system, which is needed to compute the relative goal vector $\mathbf{n}$. As shown in the supplementary video, our PI-ARS policy is able to successfully navigate to the designated target location. For the setting shown in Figure \ref{fig:navigation}, our policy successfully discovered a `shortcut' in between two obstacles and was able to go through. We do note there is collision with the obstacle's arm. This is because the robot was trained with simulated obstacles with box shapes only; further training with more diverse obstacles would likely mitigate this problem.

Overall, these experiments validate that PI-ARS is capable of learning policies that can transfer to real robots.

\section{Conclusion}
We present a new learning method, PI-ARS, and apply it to the visual-locomotion problem. PI-ARS combines gradient-based representation learning with gradient-free policy optimization to leverage the advantages of both. PI-ARS enjoys the simplicity and scalability of gradient-free methods, and it relieves a key bottleneck of ES algorithms on high-dimensional problems by simultaneously learning a low-dimensional representation that reduces the search space. We evaluate our method on a set of challenging visual-locomotion tasks, including navigating through uneven stepping stones, quincuncial piles, moving platforms, and cluttered indoor environments, among which PI-ARS significantly outperforms the state-of-the-art. Furthermore, we validate the policy learned by PI-ARS on a real quadruped robot. It enables the robot to walk over randomly-placed stepping stones and navigating in an indoor space with obstacles. In the future, we plan to test PI-ARS in outdoor visual-locomotion tasks, which presents more diverse and interesting terrains for the robot to overcome.

\section*{Acknowledgments}
We thank Noah Brown, Gus Kouretas, and Thinh Nguyen for helping set up the real-world stepping stones and address robot hardware issues. %

\bibliographystyle{IEEEtran}
\bibliography{IEEEabrv,references}

\appendix

\subsection{PI-ARS Implementation}
\label{sec:piars_implementation}

We describe implementation details of PI-ARS as follows.

\subsubsection{Network Architecture for \texorpdfstring{$\phi$}{\textit{ϕ}}}

In the visual-locomotion tasks we consider, an observation $\mathbf{s}_t$ contains a visual observation $\mathbf{s}^v_{t}$ and proprioceptive states $\mathbf{s}^p_{t}$.
The visual observation $\mathbf{s}^v_{t}$ contains one depth image from the front camera and one from the rear (described in Section~\ref{subsec:experiment_setup}).
Accordingly, our vision encoder $\phi_v$ contains two identical CNNs that independently map the front and rear images to two 64-d vectors and concatenates them into a 128-d representation output $\mathbf{z}^v_t$.
Each CNN consists of 2 convolution layers with $3\times3$ kernels, 8 channels, stride size 1, followed by a 64-d linear projection, where each convolution and projection is followed by a relu activation.
$\mathbf{z}^v_t$ is then concatenated with proprioceptive states $\mathbf{s}^p_t$ and linearly projected with tanh activation to yield a 128-d representation output $\mathbf{z}_t$.
These together give us the observation encoder $\phi$, in which $\mathbf{z}_t = \phi(\mathbf{s}^v_t, \mathbf{s}^p_t) = \phi(\mathbf{s}_t)$. 

\subsubsection{Network Architecture for Auxiliary Functions}
The auxiliary-learned function $f$ maps $\mathbf{z}_t$ to a unit-length 128-d vector $\mathbf{z}^\mathrm{past}_{t}$, which corresponds to the $\mathrm{past}$, via a 2-layer MLP (64 units, 128 units) followed by $l$-2 normalization.

For the future observation, $f$ only considers the visual observation $\mathbf{s}^v_{t+k}$ and ignores $\mathbf{s}^p_{t+k}$:
\begin{align}
\mathbf{z}^{\mathrm{future}}_{t+k} = h(\mathrm{stopgrad}(\phi_v(\mathbf{s}^v_{t+k})))
\end{align}
where $h$ is another 2-layer MLP (64 units, 128 units), and $\mathrm{stopgrad}$ refers to the stop gradient operation.
The output of $f$ is a dot-product of $\mathbf{z}^{past}_{t}$ and $\mathbf{z}^{future}_{t+k}$.

For predicting the future rewards, we recurrently apply an auxiliary RNN cell $g$ to encode a latent state and an action (i.e. the $\mathrm{past}$) and output a reward prediction and the next latent state at each time step:
\begin{align*}
\hat{r}_{t}, \mathbf{z}'_{t+1} = g(\mathbf{z}_t, \mathbf{a}_t), \hat{r}_{t+1}, \mathbf{z}'_{t+2} = g(\mathbf{z}'_{t+1}, \mathbf{a}_{t+1}), \dots,\\ 
\hat{r}_{t+k-1}, \mathbf{z}'_{t+k} = g(\mathbf{z}'_{t+k-1}, \mathbf{a}_{t+k-1}).
\end{align*}
$g$ is a 3-layer MLP (128 units each with tanh activations), and a 128-to-1 linear layer branch is attached to the second layer of $g$ to output reward predictions $\hat{r}_{t},\dots,\hat{r}_{t+k-1}$, one at each step.
The initial latent state is $\mathbf{z}_t$.

\subsubsection{Policy Network Head} The ARS-learned policy network head is a simple 3-layer MLP (64, 32, and $\mathrm{dim}(\mathcal{A})$ units, where $\mathcal{A}$ is the action space) with tanh activation that takes $\mathbf{z}_t$ and proprioceptive states $\mathbf{s}^p_t$ as input, and outputs an action.
The output of the policy network is tanh-squashed to $[-1, 1]$ and subsequently re-scaled to the environment action bounds (Table \ref{tbl:action_space}).

\subsubsection{PI-ARS Training}
In PI-ARS, we alternate between 1 step for ARS and 2 steps for PI. Each PI step uses a batch of 512 $k$-step trajectories from the replay buffer and performs gradient steps with a learning rate of $10^{-4}$.
We set $k=5$ for all tasks but indoor navigation.
For indoor navigation, we use $k=30$ as the task nature requires a longer planning horizon.
We also apply observation normalization, using running means and standard deviations.

\subsubsection{ARS, SAC, PI-SAC Implementations}
To ensure fair comparison for non-representation learning methods (ARS, SAC), we use the an identical policy network as used for PI-ARS, which combines a base encoder $\phi$ and a policy network head. 
Thus, all algorithms have access to the same capacity policy network.
The critic in SAC and PI-SAC share the same base encoder $\phi$ with the policy (actor) network and we stop gradients from the policy head to the base encoder following \cite{lee2020predictive}.
We also apply the same observation normalization used for PI-ARS to these baseline methods.

\subsection{Visual-Locomotion Task details}
\label{sec:supp_env}

Here we describe additional details regarding the visual-locomotion tasks we used in our work. For the uneven stepping stone, quincuncial piles, and moving platform tasks we follow the design in prior work \cite{yu2021visual}. 

\paragraph{Uneven stepping stones}

In this task we evaluate the ability for the policy to traverse stepping stones with varied heights. The widths, and lengths of stepping stones are sampled from $[0.55, 0.7]$, and $[0.5, 0.8]$ meters. The height offsets of neighboring stones are uniformly sampled in $[0.13, 0.2] m$, and a gap of $[0.05, 0.1] m$ is added between the stones. To successfully traverse this environment the agent needs to identify and avoid the gaps between stones and land the swing leg to the appropriate height for the next stone.

\paragraph{Quincuncial piles}

In this task, we create a two-dimensional stepping stones to evaluate the robot's behavior in avoiding gaps in both forward and lateral direction. Specifically, each stone has an area of $0.15 \times 0.15 m^2$ with a standard deviation of $0.015 m$ in height, and is separated by $[0.13, 0.17]$ m from each other in both x and y directions. 
At the beginning of each episode, we also randomly rotate entire stone grid in $[0.1, 0.1]$ rad.

\paragraph{Moving platforms}

Our proposed framework can also be applied to handle dynamic objects. In this example, we construct an environment with random stepping stones and allow each piece to move dynamically.
Each platform follows a periodic movement whose magnitude and frequency are randomly sampled in $[0.10, 0.15] m$ and $[0.4, 1.0] Hz$, respectively. 
Also, we randomly pick half of the platforms to move horizontally and the rest vertically. 
This task requires the control policy to infer both the position and velocity of the platforms and thus presents a more challenging representation learning problem. 
To enable the model to infer velocity information, we stack a history of three recent images as input to the policy for this task, which slightly changes the observation spec.

\paragraph{Indoor navigation}

For the indoor navigation task, we use a scan of the building interior with size about $25 \times 15$m. We randomly place $50$ boxes of different sizes in the scanned scene to model obstacles during navigation. Among the $50$ obstacles, $20\%$ are sampled with length in $[1.6, 2]$m, width in $[0.7, 1]$m, and height in $[0.4, 1.2]$m, which are to mimic bigger obstacles like sofa. $40\%$ are sampled with length and width in $[0.5, 0.8]$m and height in $[0.4, 1.4]$m, which represent smaller obstacles like chairs. The rest $40\%$ are sampled with length and width in $[0.1, 0.4]$m and height in $[0.8, 2.0]$m, which correspond to taller objects like pillars.

\begin{table}[t]
\caption{Action space ranges}
\label{tbl:action_space}
\begin{center}
\begin{tabular}{c|c|c}
\hline
action & lower bound & upper bound\\
\hline
Target local foothold & (-0.05, -0.05, -0.03)m & (0.1, 0.05, 0.03)m\\
Target peak swing height & 0.05m & 0.1m \\
Desired CoM height & 0.42m & 0.47m\\
Desired base roll & -0.1 & 0.1\\
Desired base pitch & -0.2 & 0.2\\
Desired base twist speed & -0.2 & 0.2\\
\hline
\end{tabular}
\end{center}
\end{table}

\end{document}